\documentclass[10pt,journal,compsoc]{IEEEtran}

\ifCLASSOPTIONcompsoc
  \usepackage[nocompress]{cite}
\else
  \usepackage{cite}
\fi

\usepackage{cite}
\usepackage{amsmath}
\usepackage{algorithmic}
\usepackage{url}
\usepackage{hyperref}
\usepackage{csquotes}
\usepackage{graphicx}
\usepackage{subfigure}
\usepackage{amssymb,bm}
\usepackage{multirow}
\usepackage{makecell}
\usepackage{multirow}
\usepackage{soul}
\graphicspath{ {./images/} }
\usepackage{xspace}
\usepackage[table]{xcolor}
\definecolor{Gray}{gray}{0.9}

\usepackage[normalem]{ulem}

\newcommand{\latinphrase}[1]{\textit{#1}} 

\newcommand{\ie}{\latinphrase{i.e.}\xspace}

\newcommand{\eg}{\latinphrase{e.g.}\xspace}
\newcommand{\etc}{\latinphrase{etc.}\xspace}
\newcommand{\wrt}{\latinphrase{w.r.t.}\xspace}

\def\ourtitle{PnP-3D: A Plug-and-Play for 3D Point Clouds}

\begin{document}

\title{\ourtitle}

\author{Shi Qiu, Saeed Anwar, and Nick Barnes
\IEEEcompsocitemizethanks{\IEEEcompsocthanksitem S. Qiu and S. Anwar are with Data61, CSIRO (The Commonwealth Scientific and Industrial Research Organisation) and School of Engineering, Australian National University, Canberra, ACT 2601, Australia.
\IEEEcompsocthanksitem N. Barnes is with School of Computing, Australian National University, Canberra, ACT 2601, Australia.
\IEEEcompsocthanksitem E-mail: \{shi.qiu, saeed.anwar, nick.barnes\}@anu.edu.au}
}


\def\NB#1{{\color{blue} {\it{#1}}}}

\IEEEtitleabstractindextext{%
\begin{abstract}
With the help of the deep learning paradigm, many point cloud networks have been invented for visual analysis. However, there is great potential for development of these networks since the given information of point cloud data has not been fully exploited. To improve the effectiveness of existing networks in analyzing point cloud data, we propose a plug-and-play module, PnP-3D, aiming to refine the fundamental point cloud feature representations by involving more local context and global bilinear response from explicit 3D space and implicit feature space. To thoroughly evaluate our approach, we conduct experiments on three standard point cloud analysis tasks, including classification, semantic segmentation, and object detection, where we select three state-of-the-art networks from each task for evaluation. Serving as a plug-and-play module, PnP-3D can significantly boost the performances of established networks. In addition to achieving state-of-the-art results on four widely used point cloud benchmarks, we present comprehensive ablation studies and visualizations to demonstrate our approach's advantages. The code will be available at \url{https://github.com/ShiQiu0419/pnp-3d}.
\end{abstract}

}

\maketitle

\IEEEdisplaynontitleabstractindextext

\IEEEpeerreviewmaketitle


\IEEEraisesectionheading{\section{Introduction}
\label{sec:intro}}
\IEEEPARstart{P}{o}int cloud data shows increasing value in both academia and industry, since affordable yet effective 3D sensors~\cite{endres20133, jaboyedoff2012use} are widely used in autonomous driving, robotics, augmented reality, \etc However, unlike the 2D images that have well-organized structure, point cloud data is \emph{unordered} and \emph{unevenly distributed} in 3D space, increasing the difficulty of machine perception towards common visual tasks such as classification, segmentation, detection, \etc

Following the success of Convolutional Neural Networks (CNNs) in 2D images, early methods~\cite{su2015multi, lawin2017deep} captured 2D multi-view projections of point clouds for 2D CNN processing. Then, to preserve the 3D context of raw data and avoid the conversion between 2D and 3D representations, researchers proposed point cloud networks such as~\cite{qi2017pointnet, qi2017pointnet++, guo2020deep} to directly process 3D data for visual analysis. However, recent research evidence suggests that the capability of existing networks has not been fully exploited: \eg, the PointAugment framework~\cite{li2020pointaugment} improves the performance of classification networks~\cite{wang2019dynamic, liu2019relation} by automatically augmenting point cloud samples; and the CGA module~\cite{lu2021cga} boosts the effectiveness of segmentation networks~\cite{hu2020randla, liu2020closer} by organizing category-aware neighborhoods. Beyond such \emph{task-specific} modules to facilitate the network's understanding of a certain task, we aim to explore a generic and effective \emph{plug-and-play} module that can be easily deployed in different point cloud networks without additional setup. To achieve broader usage in various tasks, basically, we need to emphasize a fundamental component of point cloud analysis: point feature representation.

\emph{Importance of point feature representation.} With the help of CNN's strong expression and generalization capacity, we represent point features in different dimensions of embedding space and from multiple point cloud resolutions. Then we can conduct various operations on the point feature representations to analyze some basic properties of point cloud data. To be specific, the classification task~\cite{li2020pointaugment, qiu2021geometric} usually infers the point cloud's category based on its global embedding vector, which is aggregated from the entire feature map (\ie, the feature representations of all points) via a global pooling function. Moreover, semantic segmentation networks~\cite{hu2020randla, qiu2021semantic} directly apply shared fully connected layers to each point feature representation for its semantic label prediction. Similarly, the 3D detection pipeline~\cite{qi2019deep} incorporates a point feature learning backbone generating votes for object proposals. In general, point cloud analysis significantly relies on the corresponding point feature representations; thus, for high-quality point cloud analysis, we can further \emph{refine} the learned features. Apart from the attention mechanism~\cite{vaswani2017attention} that implicitly re-weights the importance of each element in the feature map, first of all, we aim to explicitly enrich the local information of each point by incorporating different types of context.

\begin{figure*}
\begin{center}
\includegraphics[width=0.945\textwidth]{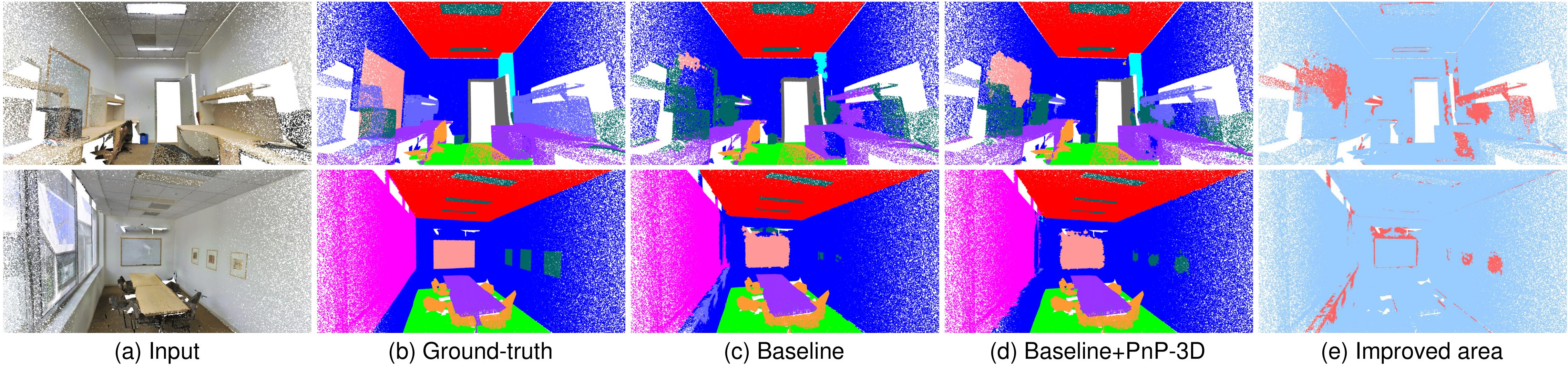}
\end{center}
\vspace{-3mm}
\caption{Examples of point cloud semantic segmentation on \emph{S3DIS}~\cite{armeni2017joint} dataset. The 3rd column shows the results of RandLA-Net~\cite{hu2020randla} (baseline), while the 4th column presents the results of plugging our PnP-3D module in the baseline. The improved areas (\ie, the points that are correctly classified by using our module but misclassified by the baseline) are highlighted with red color in the last column.}
\label{fig:vis1}
\vspace{-3mm}
\end{figure*}

\emph{Variety of point cloud context.} Recently, an increasing number of methods~\cite{yan2020pointasnl, hu2020randla, qiu2021semantic} choose to directly combine the point cloud's inherent 3D coordinates (\ie, the geometric context) with the point feature representations acquired by CNNs. The advantages can be summarized in two aspects: firstly, as the network goes deeper, the learned features in higher-dimensional embedding space will be more abstract and implicit, thus, we need to introduce the fundamental 3D geometry attribute for the network's reference; secondly, since such 3D coordinates can indicate the point distribution in the real-world, we can additionally gather more local information from the constructed point neighborhoods under a particular spatial metric such as 3D Euclidean distance in the ball query~\cite{qi2017pointnet++, liu2019relation} or the k-nearest-neighbors (knn)~\cite{wang2019dynamic, hu2020randla} algorithm. However, repeatedly and directly feeding such 3D coordinates to the network is not always an ideal solution. Instead, we try to fuse the local \emph{geometric} and \emph{feature} context, which are captured in parallel following the same constraints and formulation derived from the point cloud's inherent 3D coordinates. By gathering the local information, not only do we introduce 3D geometric relationships for the network, but we also provide more local distinctness in point feature representation.

{\emph{Global perception of point cloud data.}} Besides incorporating different types of local context, we still need to refine point feature representations from a global perspective (\wrt the entire feature map) for two main purposes: (i) to strengthen the inter-dependencies between point-wise and channel-wise elements; (ii) to regularize all the learned point features for the machine's explicit understanding. Although some previous methods~\cite{xie2018attentional, yan2020pointasnl, qiu2021geometric} employed self-attention structures~\cite{wang2018non} to estimate element-wise dependencies, due to the complexity of point cloud data and networks, extra heavy computations for self-attention are not practically preferable. Alternatively, we can use lightweight operations to generate a global bilinear response consisting of both point-wise and channel-wise inter-dependencies. Coupled with rich local information, a comprehensive feature map can be synthesized for accurate point cloud analysis.

\section{Related Work}
\label{sec:work}
\noindent \textbf{Point Cloud Analysis Tasks.}
As a fundamental problem in point cloud analysis, the classification task aims to identify the category of a point cloud \emph{object}. To deal with the unordered points, regular methods~\cite{qi2017pointnet, qiu2021geometric, li2020pointaugment} apply a symmetric function (\eg, max-pooling) to aggregate a global embedding from its feature map. Based on the global embedding, we further regress the possibilities regarding candidate categories via fully connected layers. For a more fine-grained analysis of a point cloud \emph{scene}, we can segment the whole point cloud into a few clusters based on different types of point-related properties such as the object-part class~\cite{lyu2020learning, qiu2021dense}, semantic category~\cite{hu2020randla, qiu2021semantic}, or instance index~\cite{han2020occuseg, jiang2020pointgroup}. Particularly, the basic point cloud semantic segmentation relies highly on the point feature representation to predict each point's semantic label. Moreover, as autonomous driving techniques develop rapidly, point cloud object detection is becoming more popular in the 3D computer vision area. As the main track of solutions, the region proposal-based methods~\cite{qi2019deep, qi2020imvotenet, xie2020mlcvnet} utilize a backbone network learning the point feature representations and predict the 3D bounding boxes and categories of the physical objects based on the learned information. 

To conclude, point feature representations are crucial in point cloud analysis. By using our \emph{plug-and-play} module to synthesize a more comprehensive feature map, the effectiveness of point cloud networks can be further improved.

\vspace{2mm}
\noindent \textbf{Point-based Networks.}
Unlike projection-based~\cite{su2015multi, lawin2017deep} or discretization-based~\cite{choy20194d, su2018splatnet} networks which \emph{indirectly} analyze point cloud data, point-based networks~\cite{qi2017pointnet, qi2017pointnet++} are more widely adopted in point cloud analysis because of the intuitiveness and simplicity. To be specific, the basic point-based networks usually apply the Multi-Layer-Perceptron (MLP)~\cite{qi2017pointnet} to \emph{directly} learn the point-wise features from 3D point cloud data, while the advanced methods~\cite{qi2017pointnet++, engelmann2020dilated} tend to form point neighborhoods for more local context. Later works extend the primary usage of MLPs in different ways: the graph-based methods~\cite{wang2019dynamic, qiu2021dense} extract local features from crafted point graphs; and the convolution-based approaches~\cite{xu2018spidercnn, liu2019relation, thomas2019kpconv} develop several MLP variants by involving more geometric clues.

Even though most point-based networks learn the point feature representations from multiple point cloud resolutions, such a \emph{plug-and-play} module can be easily and flexibly deployed to refine the feature map of each resolution, showing great adaptability in point-based networks.

\begin{figure*}
\begin{center}
\includegraphics[width=0.97\textwidth]{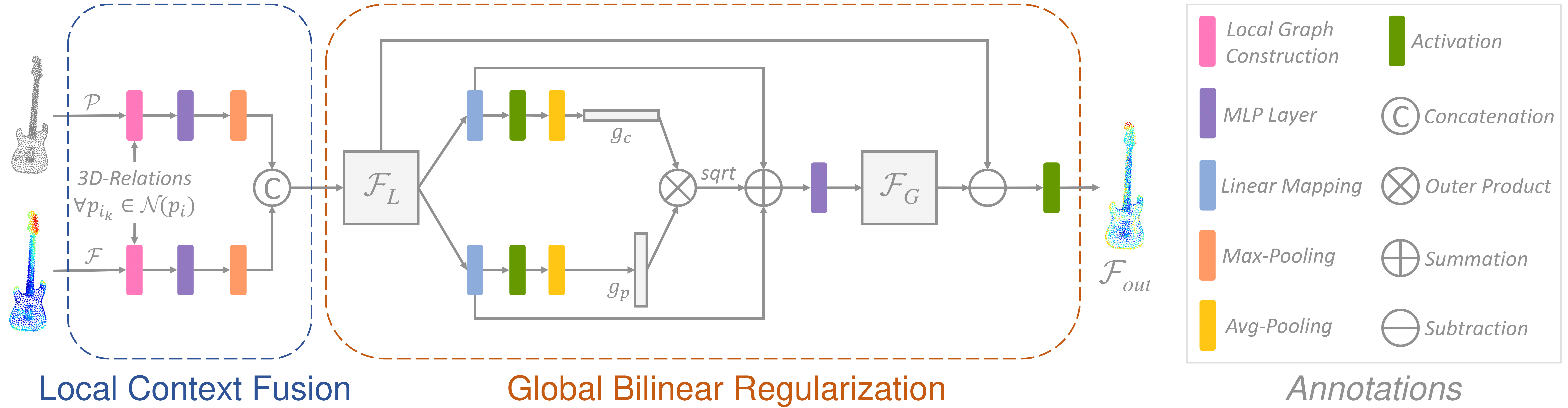}
\end{center}
\vspace{-3mm}
\caption{Detailed structure of the PnP-3D module. The local context fusion block (Section~\ref{sec:local}) aims to fuse local geometric and feature context based on the geometric relations between points in 3D space. Further, the global bilinear regularization block (Section~\ref{sec:global}) can regularize the feature map by aggregating global perceptions based on both point-wise and channel-wise information in feature space. More implementation details are in Section~\ref{sec:imp}.}
\label{fig:module}
\vspace{-3mm}
\end{figure*}

\vspace{2mm}
\noindent \textbf{Attention Mechanism for Point Clouds.}
As a powerful tool in deep learning, attention mechanisms~\cite{vaswani2017attention} have been utilized to address many computer vision problems. Similar to the Non-local network~\cite{wang2018non} and SENet~\cite{hu2018squeeze} for 2D images, we are able to re-weight the point cloud feature map by applying attention-based structures along either point-axis or channel-axis. To be concrete,~\cite{xie2018attentional} and~\cite{feng2020point} follow the self-attention pipeline to calculate the long-range dependencies between points, while GBNet~\cite{qiu2021geometric} regresses the channel-wise affinities in a similar behavior. In addition,~\cite{wang2019graph} and~\cite{chen2021gapointnet} leverage the idea of graph-attention~\cite{velivckovic2017graph} in graph-based point cloud networks. More recently,~\cite{zhao2020point} and~\cite{guo2021pct} achieve good performance on the point cloud classification and segmentation benchmarks by forming a transformer-based network~\cite{khan2021transformers}, where the regular point feature encoders and decoders are replaced with the attention-based modules. 

Given the complexity of point cloud data, we propose an effective \emph{plug-and-play} module, exceeding the function of regular attention approaches in two ways: (i) more local information is incorporated to enrich the point context; and (ii) both point-wise and channel-wise global perceptions are exploited to regulate the feature representations.
\section{Approach}
\label{sec:app}
\subsection{Overview}
Given a point cloud with $N$ points, the 3D coordinates can be presented as $\mathcal{P}=[p_1;...;p_i;...;p_N]\in\mathbb{R}^{N\times3}$, where an arbitrary point is denoted as ${p_i}\in\mathbb{R}^{3}$. Moreover, we can use CNN-based modules to learn the feature map $\mathcal{F}=[f_1;...;f_i;...;f_N]\in\mathbb{R}^{N\times C}$ in a $C$-dimensional space, where $p_i$'s corresponding feature representation is ${f_i}\in\mathbb{R}^{C}$.

In language-related usage~\cite{vaswani2017attention} of the attention mechanism, the positional encoding operation gives a cue for \emph{sequence order} to the network by inputting the position of tokens. Similarly, in the case of unordered point cloud data, we need to inform the network about the \emph{geometric relations} between scattered points. As shown in Figure~\ref{fig:module}, we deploy a local context fusion block to fulfill this intention.

\subsection{Local Context Fusion}
\label{sec:local}
Following either a knn~\cite{wang2019dynamic} or ball-query algorithm~\cite{qi2017pointnet++}, we can simply find the neighbors ${\forall}p_{i_k}\in \mathcal{N}(p_i)$ of a certain point $p_i$, under a metric of \emph{3D Euclidean distance} between the scattered points. By combining the edges~\cite{wang2019dynamic} between $p_i$ itself and its $k$ neighbors ${\forall}p_{i_k}\in \mathcal{N}(p_i)$, we define a local geometric graph of $p_i$ in 3D space follows: $\tilde{p_i} = [p_i; p_{i_k}-p_i]$, $\tilde{p_i}\in\mathbb{R}^{k\times 6}$. Thus, the local geometric graphs for all points $\mathcal{P}$ are generally denoted as $\tilde{\mathcal{P}} = [\tilde{p_1};...;\tilde{p_i};...;\tilde{p_N}] \in\mathbb{R}^{N\times k \times 6}$.

Further, we encode local geometric graphs via a shared MLP $\bm{\mathcal{M}}_{\Theta}$, and aggregate the local geometric context by applying the max-pooling function over the $k$ neighbors:
\begin{equation}
\label{equ:local}
    \mathring{\mathcal{P}} = \max_{k}\big(\bm{\mathcal{M}}_{\Theta}(\tilde{\mathcal{P}})\big), \quad \mathring{\mathcal{P}}\in\mathbb{R}^{N\times \frac{C}{2}};
\end{equation}
practically we implement the shared MLP operation as a $1\times 1$ convolution followed by a batch normalization (BN) and an activation layer. 

In parallel, a local feature graph of $f_i$ in $C$-dimensional space can be formed as: $\tilde{f_i} = [f_i; f_{i_k}-f_i] \in\mathbb{R}^{k\times 2C}$, where $\forall f_{i_k}$ are corresponding features of ${\forall}p_{i_k}\in \mathcal{N}(p_i)$. Accordingly, local feature graphs of the feature map $\mathcal{F}$ are represented as: $\tilde{\mathcal{F}} = [\tilde{f_1};...;\tilde{f_i};...;\tilde{f_N}] \in\mathbb{R}^{N\times k \times 2C}$. Following similar operations in Equation~\ref{equ:local}, we obtain the local feature context:
\begin{equation}
    \mathring{\mathcal{F}} = \max_{k}\big(\bm{\mathcal{M}}_{\Phi}(\tilde{\mathcal{F}})\big), \quad \mathring{\mathcal{F}}\in\mathbb{R}^{N\times \frac{C}{2}};
\end{equation}
where $\bm{\mathcal{M}}_{\Phi}$ is another shared MLP encoding local feature graphs. Finally, we concatenate the local geometric context $\mathring{\mathcal{P}}$ and the local feature context $\mathring{\mathcal{F}}$ as the output of our local context fusion block: 
\begin{equation}
    {\mathcal{F}_L}= \mathrm{concat}(\mathring{\mathcal{P}}, \mathring{\mathcal{F}}), \quad {\mathcal{F}_L}\in\mathbb{R}^{N\times C}.
\end{equation}

Compared to the \emph{EdgeConv}~\cite{wang2019dynamic} operation, both the local geometric and feature context in our approach are formulated and constrained with inherent 3D geometric relations. Furthermore, the local context fusion block can be flexibly deployed at different point cloud resolutions and CNN layers, benefiting most of the existing point cloud networks.

\subsection{Global Bilinear Regularization}
\label{sec:global}
In addition to gathering more local detail for each point's feature representation, the global bilinear regularization block aims to refine the feature map by taking global perceptions of the whole point cloud into account. Recall that in regular self-attention mechanisms, global perceptions are estimated as long-range dependencies (\ie, cosine similarities) between point-wise features, consuming a relatively large amount of memory. Instead, global perceptions in our approach are computed as the feature map's element-level inter-dependencies based on the global channel-wise and point-wise descriptors.  

To encode the global channel-wise descriptor: first, we apply a weight matrix, $\bm{W_c}\in\mathbb{R}^{C\times \frac{C}{r}}$ where $r$ is a reduction factor, to reduce the dimension of the fusion output ${\mathcal{F}_L}\in\mathbb{R}^{N\times C}$; then, we leverage the $\mathrm{ReLU}$~\cite{nair2010rectified} function to not only provide non-linearity after the linear mapping with $\bm{W_c}$, but also meet the demand of non-negativity in Equation~\ref{equ:g_mean}; finally, by conducting the average-pooling operation over the $N$ elements along space-axis, we can squeeze the spatial information~\cite{hu2018squeeze} as a global channel-wise descriptor $g_c$. Mathematically, the above operations follow:
\begin{equation}
\label{equ:global_c}
    g_c = \underset{N}{\mathrm{avg}}\big(\mathrm{ReLU}({\mathcal{F}_L}\bm{W_c})\big), \quad g_c\in\mathbb{R}^{\frac{C}{r}};
\end{equation}
where \enquote{${\mathcal{F}_L}\bm{W_c}$} is the \emph{matrix product} between ${\mathcal{F}_L}$ and $\bm{W_c}$, \enquote{${\mathrm{avg}}$} is the average-pooling, and the reduction factor (an integer) satisfies: $r\geq 2$. Further, $g_c = [{\mu}_1,...,{\mu}_j,...,{\mu}_{\frac{C}{r}}]$, and ${\mu}_j\in\mathbb{R}$ stands for the global (mean) response of the $j$-th \emph{channel} in a feature map of whole point cloud.

With another weight matrix $\bm{W_p}\in\mathbb{R}^{C\times \frac{C}{r}}$, ReLU function, and the average-pooling operation over the $\frac{C}{r}$ elements along channel-axis, we can also generate a global point-wise descriptor $g_p$ in a similar way to that used in Equation~\ref{equ:global_c}: 
\begin{equation}
\label{equ:global_p}
    g_p = \underset{\frac{C}{r}}{\mathrm{avg}}\big(\mathrm{ReLU}({\mathcal{F}_L}\bm{W_p})\big), \quad g_p\in\mathbb{R}^{N},
\end{equation}
where $g_p = [{\lambda}_1,...,{\lambda}_i,...,{\lambda}_N]$ and ${\lambda}_i\in\mathbb{R}$ estimate the global (mean) response of the $i$-th \emph{point} in the whole point cloud's feature map; \enquote{${\mathcal{F}_L}\bm{W_p}$} is another \emph{matrix product}.

Unlike~\cite{kim2017hadamard} which uses a Hadamard product (\ie, element-wise product) between the vectors, we capture a low-rank global bilinear response by taking the square root of $g_p$ and $g_c$'s \emph{outer} product:
\begin{equation}
\label{equ:g_mean}
    \mathcal{G} = \mathrm{sqrt} (g_p \otimes g_c), \quad \mathcal{G}\in\mathbb{R}^{N\times \frac{C}{r}};
\end{equation}
where an element $\eta_{ij}$ positioned at the $i$-th row and $j$-th column of $\mathcal{G}$ is mathematically calculated as:
\begin{equation}
    \eta_{ij} = \sqrt{{\lambda}_i {\mu}_j}, \quad \eta_{ij}\in\mathbb{R}.
\end{equation}
The reasons of using a global bilinear response can be explained from two aspects. Firstly, given that a global channel-wise descriptor captures channel-wise dependencies~\cite{hu2018squeeze} and a global point-wise descriptor represents overall shape context~\cite{qi2017pointnet}, calculating a bilinear combination of these two global descriptors can comprehensively synthesize and fully preserve the corresponding two types of global information. Secondly, in terms of each element, as $\lambda_i$ and $\mu_j$ are the \emph{arithmetic} means of the $i$-th point and $j$-th channel respectively, $\eta_{ij}$ is the \emph{geometric mean} of $\lambda_i$ and $\mu_j$: it provides a higher-order mean response based on both spatial and channel-related information.

After applying a shared MLP $\bm{\mathcal{M}}_{\Psi}$ and two shortcut connections as suggested in~\cite{kim2017hadamard}, we finally restore the channel dimension, and generate a full-sized global perception map:
\begin{equation}
    \mathcal{F}_G = \bm{\mathcal{M}}_{\Psi}\big(\mathcal{G} + {\mathcal{F}_L}\bm{W_c} + {\mathcal{F}_L}\bm{W_p}\big); 
\end{equation}
where $\mathcal{F}_G\in\mathbb{R}^{N\times C}$. Next, we expect to find a better usage of the global perception map. 

The average-pooling operation in Equations~\ref{equ:global_c} and~\ref{equ:global_p} is used to squeeze the global information from the point space and channel space respectively, where the pooled mean values usually represent the \emph{common} patterns~\cite{boureau2010learning, lin2013network} in feature maps. However, in point cloud analysis, we would like the learned features to be more distinct and representative. In this case, to sharpen the learned features, the global perception ${\mathcal{F}_G}$ that is derived from two global mean vectors, $g_p$ and $g_c$, can be further filtered out. In practice, we achieve this by subtracting $\mathcal{F}_G$ from the local context fusion output $\mathcal{F}_L$, and use an activation $\sigma$ to add more non-linearity in the final output feature map:
\begin{equation}
\label{equ:out}
    \mathcal{F}_{out} = \sigma(\mathcal{F}_{L} - \mathcal{F}_{G}), \quad \mathcal{F}_{out}\in\mathbb{R}^{N\times C}. 
\end{equation}
As the selection of pooling and regularization strategy is always an open but practical question, we conduct the related ablation studies to investigate the best form of our global bilinear regularization block. More details and discussions can be found in Section~\ref{sec:abl}. 
\section{Implementation Details}
\label{sec:imp}
Since we mainly propose a \emph{plug-and-play} module for point cloud analysis networks, for fair comparisons, most of the hyperparameters (\eg, feature dimensions, training paradigms, pre/post-processing \etc) in our experiments are adopted from the baseline's implementation if not explicitly mentioned. In general, the PnP-3D module would be placed after each feature encoding layer (encoder) of a baseline network; more concrete settings are provided in Section~\ref{sec:exp}.

As for the local context fusion block, the neighbors are searched and collected in the same way as in the baseline network. Particularly, in the point cloud semantic segmentation task, we only take half of the original $k$ neighbors as suggested in~\cite{lu2021cga} to save memory. Further, to minimize the module's complexity, all shared MLPs are implemented as the composition of a \emph{single-layer} $1\times 1$ convolution, a BN layer, and a $\mathrm{ReLU}$ activation.  

In terms of the global bilinear regularization block, the matrix multiplication operation in Equation~\ref{equ:global_c} and~\ref{equ:global_p} is realized as a \emph{single-layer} $1\times 1$ convolution, which can linearly map the features into a lower-dimensional space. Empirically, we set the reduction factor $r=8$ in all cases. Moreover, a self-regularized and non-monotonic activation function, Mish~\cite{misra2019mish}, is leveraged as $\sigma$ in Equation~\ref{equ:out}.

\begin{table}
\caption{Overall classification results (\%) on \emph{ModelNet40}~\cite{wu20153d} and \emph{ScanObjectNN}~\cite{Uy_2019_ICCV}. (\enquote{OA}: Overall Accuracy; \enquote{mAcc}: mean Class Accuracy; {\textbf{Bold}} numbers indicate the results higher than corresponding baselines. For each column, the \emph{highest} value is highlighted in \textcolor{red}{red}.)}
\vspace{-3mm}
\begin{center}
\resizebox{0.8\columnwidth}{!}{
\begin{tabular}{|c|c|c|c|c|}\hline
\multirow{2}{*}{Method} & \multicolumn{2}{c|}{\emph{ModelNet40}} & \multicolumn{2}{c|}{\emph{ScanObjectNN}} \\ \cline{2-5} 
& OA            & mAcc            & OA             & mAcc             \\ \hline\hline
PointNet~\cite{qi2017pointnet} &89.2 &86.0 &68.2 &63.4 \\ \hline
PointCNN~\cite{li2018pointcnn} &92.2 &88.1 &78.5 &75.1\\ \hline
SpiderCNN~\cite{xu2018spidercnn} &92.4 &- &73.7 &69.8\\ \hline
DRNet~\cite{qiu2021dense} &93.1 &- &80.3 &78.0\\ \hline
RS-CNN~\cite{liu2019relation} &92.2 &88.3 &75.2 &71.6\\
\rowcolor{Gray}\textbf{+PnP-3D} &{\textbf{93.1}} &\textbf{89.5} &\textbf{77.9} &\textbf{73.6}\\
\hline
PointNet++~\cite{qi2017pointnet++} &90.7 &87.6 &77.9 &75.4\\
\rowcolor{Gray}\textbf{+PnP-3D} &{\textbf{93.2}} &\textbf{91.1}&\textcolor{red}{\textbf{82.2}} &\textcolor{red}{\textbf{79.6}}\\
\hline
DGCNN~\cite{wang2019dynamic} &92.2 &90.2 &78.1 &73.6\\
\rowcolor{Gray}\textbf{+PnP-3D} &\textbf{93.4} &{\textbf{90.8}} &{\textbf{81.0}} &{\textbf{78.0}}\\\hline
PAConv~\cite{xu2021paconv} &93.6 &- &77.1 &74.4\\
\rowcolor{Gray}\textbf{+PnP-3D} &\textcolor{red}{\textbf{93.8}} &\textcolor{red}{{\textbf{91.4}}} &{\textbf{80.3}} &{\textbf{76.8}}\\\hline
AdaptConv~\cite{zhou2021adaptive} &93.4 &90.7 &79.3 &76.0\\
\rowcolor{Gray}\textbf{+PnP-3D} &\textcolor{red}{\textbf{93.8}} &{\textbf{91.1}} &{\textbf{81.5}} &{\textbf{78.9}}\\
\hline
\end{tabular}
}
\end{center}
\label{tab:cls}
\vspace{-3mm}
\end{table}

\begin{table*}
\caption{Detailed semantic segmentation results (Intersection-over-Union, \%) on the \emph{Area 5} of \emph{S3DIS}~\cite{armeni2017joint} dataset. (\enquote{mIoU}: mean Intersection-over-Union; {\textbf{Bold}} numbers indicate the results higher than corresponding baselines. For each column, the \emph{highest} value is highlighted in \textcolor{red}{red}.)}
\vspace{-3mm}
\begin{center}
\resizebox{0.95\textwidth}{!}{
\begin{tabular}{c|c| c c c c c c c c c c c c c}
\Xhline{3\arrayrulewidth}
Method &mIoU&ceiling&floor&wall&beam&column&window&door&table&chair&sofa&bookcase&board&clutter\\\hline\hline
PointNet~\cite{qi2017pointnet} &41.1 &88.8 &97.3 &69.8 &\textcolor{red}{0.1} &3.9 &46.3 &10.8 &52.6 &58.9 &40.3 &5.9 &26.4 &33.2\\
PointCNN~\cite{li2018pointcnn} &57.3 &92.3 &98.2 &79.4 &0.0 &17.6 &22.8 &62.1 &74.4 &80.6 &31.7 &66.7 &62.1 &56.7\\
SPGraph~\cite{landrieu2018large} &58.0 &89.4 &96.9 &78.1 &0.0 &\textcolor{red}{42.8} &48.9 &61.6 &84.7 &75.4 &69.8 &52.6 &2.1 &52.2\\
PointWeb~\cite{zhao2019pointweb} &60.3 &92.0 &{98.5} &79.4 &0.0 &21.1 &59.7 &34.8 &76.3 &{88.3} &46.9 &69.3 &64.9 &52.5\\
PointASNL~\cite{yan2020pointasnl} &62.6 &{94.3} &98.4 &79.1 &0.0 &26.7 &55.2 &66.2 &83.3 &86.8 &47.6 &68.3 &56.4 &52.1\\
Minkowski-Net~\cite{choy20194d} &65.4 &91.8 &\textcolor{red}{98.7} &\textcolor{red}{86.2} &0.0 &34.1 &48.9 &62.4 &81.6 &\textcolor{red}{89.8} &47.2 &{74.9} &\textcolor{red}{74.4} &58.6\\
BAAF-Net~\cite{qiu2021semantic} &65.4 &92.9 &97.9 &82.3 &0.0 &23.1 &\textcolor{red}{65.5} &64.9 &78.5 &87.5 &61.4 &70.7 &68.7 &57.2\\
KPConv~\cite{thomas2019kpconv} &67.1 &92.6 &97.3 &81.4 &0.0 &16.5 &54.5 &{69.5} &{90.1} &80.2 &74.6 &66.4 &63.7 &58.1\\
JSENet~\cite{hu2020jsenet} &{67.7} &93.8 &97.0 &83.0 &0.0 &23.2 &61.3 &\textcolor{red}{71.6} &89.9 &79.8 &75.6 &72.3 &{72.7} &{60.4}\\
\hline
PointNet++~\cite{qi2017pointnet++} & 53.5 &89.4&97.7&75.4&0.0&1.8&58.3&19.5&69.2&79.0&46.2&59.1&58.7&41.6\\
\rowcolor{Gray}\textbf{+PnP-3D} & \textbf{58.7} &\textbf{90.8}&\textbf{98.1}&\textbf{77.0}&0.0&\textbf{7.6}&\textbf{61.7}&\textbf{28.1}&\textbf{74.2}&\textbf{84.3}&\textbf{67.3}&\textbf{64.8}&\textbf{60.5}&\textbf{48.0}\\ \hline
RandLA-Net~\cite{hu2020randla} &62.5 &92.3 &97.7 &80.5 &0.0 &20.9 &62.0 &35.3 &77.7 &86.8 &74.7 &68.8 &65.0 &50.8\\
\rowcolor{Gray}\textbf{+PnP-3D} &\textbf{65.7} &\textbf{92.5}&97.6&\textbf{{82.0}}&0.0&{\textbf{{34.4}}}&\textbf{{64.0}}&\textbf{{52.1}}&\textbf{{78.5}}&86.8&{\textbf{{75.4}}}&\textbf{{70.0}}&\textbf{{69.5}}&\textbf{{51.8}}\\ \hline
SCF-Net~\cite{fan2021scf} & 63.1 &90.4 &97.0 &80.6  &0.0 &19.3 &57.2 &44.8 &78.6 &86.9 &75.9 &71.4 &68.4 &49.9\\
\rowcolor{Gray}\textbf{+PnP-3D} &\textbf{65.8} &\textbf{91.2} &\textbf{97.1} &\textbf{82.0}  &0.0 &\textbf{26.9} &\textbf{62.4} &\textbf{51.9} &78.6 &\textbf{88.6} &\textcolor{red}{\textbf{80.2}} &\textbf{71.6} &\textbf{73.0} &\textbf{51.4}\\\hline
CloserLook3D~\cite{liu2020closer} & 65.7 &93.9 &98.3 &82.0 &0.0 &18.2 &56.6 &68.0 &91.2 &80.3 &75.3 &58.4 &70.6 &60.8\\
\rowcolor{Gray}\textbf{+PnP-3D} & \textcolor{red}{\textbf{68.5}} &\textcolor{red}{\textbf{94.7}} &\textbf{98.4} &{\textbf{83.7}} &0.0 &\textbf{21.1} &\textbf{60.4} &64.4 &\textcolor{red}{\textbf{92.9}} &\textbf{83.1} &\textbf{76.8} &\textcolor{red}{\textbf{83.5}} &69.1 &\textcolor{red}{\textbf{62.2}}\\ 
\Xhline{3\arrayrulewidth}
\end{tabular}
}
\end{center}
\label{tab:s3dis}
\vspace{-3mm}
\end{table*}

\section{Experiments}
\label{sec:exp}
To comprehensively validate the effectiveness of our approach in point cloud analysis, we conduct a wide range of experiments, including point cloud classification, semantic segmentation, and object detection using different baselines and datasets. In each of the following subsections, we provide the experimental details for a specific task. 

\subsection{Point Cloud Classification}
\label{sec:cls}
\noindent \textbf{Classification Baselines.}
As an extension of PointNet~\cite{qi2017pointnet}, PointNet++~\cite{qi2017pointnet++} additionally encodes each point feature from a ball-like local area found by the ball-query algorithm. A U-Net shape investigates lower point cloud resolutions and then higher resolutions. In a similar architecture, RS-CNN~\cite{liu2019relation} replaces the regular MLP with the relation shape convolution (RS-Conv) in PointNet++. On the contrary, DGCNN~\cite{wang2019dynamic} gradually represents the full-sized point cloud feature maps using cascaded encoders, by which the point-level graphs can be dynamically crafted. Moreover, the latest PAConv~\cite{xu2021paconv} and AdaptConv~\cite{zhou2021adaptive} networks are also tested.

\vspace{2mm}
\noindent \textbf{Classification Dataset.}
ModelNet40~\cite{wu20153d} is a synthetic dataset made up of 12,311 CAD-generated point cloud samples in 40 different object categories. In particular, the corresponding point cloud data is uniformly sampled from the surface of each mesh. Following the official data split, we have 9,843 point clouds in the training set, while the remaining 2,468 point clouds are for testing. In addition, ScanObjectNN~\cite{Uy_2019_ICCV} has around 14,298 point clouds (11,416 training samples and 2,882 testing samples) in 15 classes, which are manually scanned from a real-world environment. As suggest in~\cite{Uy_2019_ICCV}, we use the hardest perturbation variant of ScanObjectNN for our experiments.

\vspace{2mm}
\noindent \textbf{Experimental Settings.}
In general, we take the 3D coordinates of 1024 points as each point cloud's input for all classification experiments. Specifically, for both RS-CNN and PointNet++, the single-scale model is deployed as our baseline, where the PnP-3D module is placed after each Set Abstraction (SA) layer. For the rest baselines, we apply our module to refine each output of the convolution block.

\vspace{2mm}
\noindent \textbf{Classification results.}
Table~\ref{tab:cls} presents the detailed classification results on ModelNet40 and ScanObjectNN datasets. Using the PnP-3D module, all five tested baseline networks achieve state-of-the-art performances (over 93\% in overall accuracy) on ModelNet40 under the most basic input condition (3D coordinates of 1024 points). Further, after deploying the PnP-3D module, all baselines obtain significant improvements (2.2\% to 4.3\%) on ScanObjectNN compared with their original results. Particularly, as the most widely used baseline, PointNet++ shows great potential with the help of our approach beating all methods listed on ScanObjectNN leaderboard\footnote{\url{https://hkust-vgd.github.io/scanobjectnn/}}.   

\subsection{Point Cloud Semantic Segmentation}
\label{sec:seg}
\noindent \textbf{Semantic Segmentation Baselines.}
In order to verify the effectiveness of our module on the point cloud semantic segmentation task, besides PointNet++, we adopt another three recently introduced networks. Specifically, both RandLA-Net~\cite{hu2020randla} and SCF-Net~\cite{fan2021scf} take advantage of Random Sampling and Nearest Neighbor Interpolation methods to boost the efficiency in processing large-scale point cloud data, while CloserLook3D~\cite{liu2020closer} involves different local aggregation operators in a typical ResNet~\cite{he2016deep} architecture.

\vspace{2mm}
\noindent \textbf{Semantic Segmentation Dataset.}
S3DIS~\cite{armeni2017joint} is precisely scanned from 6 main indoor working areas containing a total of 272 rooms in different types such as office, storage or conference room, \etc In particular, each room's point cloud data is made up of 0.5 to 2.5 million points labeled in 13 semantic classes. In the experiment, we take Area 5 as the test set while the remaining five areas are reserved for training. For fair comparisons, the RGB colors of points are used as additional input information. 

\begin{table*}
\caption{Detailed object detection results (Average Precision, \%) on the \emph{validation set} of \emph{SUN RGB-D V1}~\cite{song2015sun} dataset. (\enquote{mAP}: mean AP, IoU threshold of 0.25~\cite{song2015sun}; {\textbf{Bold}} numbers indicate the results higher than corresponding baselines. For each column, the \emph{highest} value is highlighted in \textcolor{red}{red}.)}
\vspace{-3mm}
\begin{center}
\resizebox{0.82\textwidth}{!}{
\begin{tabular}{c|c| c c c c c c c c c c}
\Xhline{3\arrayrulewidth}
\multirow{2}{*}{Method}&\multirow{2}{*}{mAP} &bath&\multirow{2}{*}{bed}&book&\multirow{2}{*}{chair}&\multirow{2}{*}{desk}&\multirow{2}{*}{dresser}&night&\multirow{2}{*}{sofa}&\multirow{2}{*}{table}&\multirow{2}{*}{toilet} \\
& &tub&&shelf&&&&stand&&&\\\hline \hline
DSS~\cite{song2016deep} &42.1 &44.2 &78.8 &11.9 &61.2 &20.5 &6.4 &15.4 &53.5 &50.3 &78.9\\
COG~\cite{ren2016three} &47.6 &58.3 &63.7 &31.8 &62.2 &\textcolor{red}{45.2} &15.5 &27.4 &51.0 &{51.3} &70.1\\
2D-driven~\cite{lahoud20172d} &45.1 &43.5 &64.5 &31.4 &48.3 &27.9 &25.9 &41.9 &50.4 &37.0 &80.4\\
F-PointNet~\cite{qi2018frustum} &54.0 &43.3 &81.1 &33.3 &64.2 &24.7 &32.0 &58.1 &61.1 &51.1 &90.9 \\
MLCVNet~\cite{xie2020mlcvnet} &59.8 &{79.2} &85.8 &31.9 &75.8 &26.5 &31.3 &61.5 &66.3 &50.4 &89.1\\
MLCVNet++~\cite{xie2021vote} &60.9 &\textcolor{red}{79.3} &85.3 &36.5 &\textcolor{red}{77.1} &28.7 &31.6 &61.4 &68.3 &50.7 &90.0\\
\hline
BoxNet~\cite{qi2019deep} & 52.4 &63.4 &85.4 &33.4 &70.9 &20.2 &22.6 &38.5 &65.8 &41.1 &83.3\\
\rowcolor{Gray}\textbf{+PnP-3D} &\textbf{54.1}&\textbf{73.5}&83.7&\textbf{34.3}&69.9&\textbf{20.6}&21.2&\textbf{44.2}&63.0&\textbf{41.6}&\textbf{88.6}\\ \hline
VoteNet~\cite{qi2019deep} & 57.7 &74.4 &83.0 &28.8 &75.3 &22.0 &29.8 &62.2 &64.0 &47.3 &90.1\\
\rowcolor{Gray}\textbf{+PnP-3D} & \textbf{59.1} &\textbf{74.8} &\textbf{84.7} &\textbf{32.1} &75.0 &\textbf{26.4} &28.8 &\textbf{66.0} &\textbf{64.2} &\textbf{49.3} &89.4\\ \hline
ImVoteNet~\cite{qi2020imvotenet} &{62.3} &70.5 &{87.8} &{41.9} &75.6 &27.6 &{39.9} &{67.0} &\textcolor{red}{70.7} &50.2 &{91.6}\\
\rowcolor{Gray}\textbf{+PnP-3D} &\textcolor{red}{\textbf{63.8}} &\textbf{74.9} &\textcolor{red}{\textbf{88.6}} &\textcolor{red}{\textbf{42.9}} &{\textbf{76.3}} &{\textbf{28.1}} &\textcolor{red}{\textbf{41.7}} &\textcolor{red}{\textbf{69.5}} &{70.5} &\textcolor{red}{\textbf{52.2}} &\textcolor{red}{\textbf{92.7}}\\ 
\Xhline{3\arrayrulewidth}
\end{tabular}
}
\end{center}
\label{tab:sunrgbd}
\vspace{-3mm}
\end{table*}

\vspace{2mm}
\noindent \textbf{Experimental Settings.}
Similarly, as in Section~\ref{sec:cls}, the PnP-3D module is placed behind the SA layer in single-scale PointNet++ and the encoder in RandLA-Net and SCF-Net. As for CloserLook3D, our module refines the output feature map in every stage of its backbone (width=144, repeating factor=1, bottleneck ratio=2). Additionally, to testify to our module's effects on refining \emph{pseudo grid} based features that are heavily exploited in~\cite{li2018pointcnn, thomas2019kpconv, choy20194d}, we use the \emph{pseudo grid} local aggregation operator in CloserLook3D baseline. 

\vspace{2mm}
\noindent \textbf{Semantic Segmentation results.}
As shown in Table~\ref{tab:s3dis}, the PnP-3D module can significantly boost the baseline's overall performances (by 2.7\% to 5.2\% mIoU) on the S3DIS dataset. Most of the 13 semantic categories can achieve varying degrees of improvement for each baseline by using our module. In terms of the categories like \emph{column, door, sofa, and bookcase}, the corresponding IoUs show substantial growth in all networks, where the highest growth is more than 25\%. In particular, we successfully achieve the highest mIoU, 68.5\%, on Area 5 test set by utilizing the PnP-3D module in the CloserLook3D baseline. 

\begin{table}
\begin{center}
\caption{Selecting the pooling and regularization strategy, tested on RandLA-Net~\cite{hu2020randla} and \emph{Area 5} of \emph{S3DIS}~\cite{armeni2017joint} dataset. (\enquote{$\odot$}, \enquote{$\oplus$}, \enquote{$\ominus$}: element-wise product/summation/subtraction.)}
\vspace{-3mm}
\resizebox{0.9\columnwidth}{!}{
\begin{tabular}{c|c|c|c|c}
\Xhline{3\arrayrulewidth}
\multirow{2}{*}{Model} &Operation&Operation&Regularization &\multirow{2}{*}{mIoU}\\
&(in Equation~\ref{equ:global_c})&(in Equation~\ref{equ:global_p})&(in Equation~\ref{equ:out}) &\\\hline
1 &max-pooling &max-pooling & $\odot$&63.0\\
2 &max-pooling &max-pooling & $\oplus$&64.3\\
3 &max-pooling &max-pooling & $\ominus$&63.3\\
4 &avg-pooling &avg-pooling & $\odot$&63.7\\
5 &avg-pooling &avg-pooling & $\oplus$&64.5\\
\hline
\textbf{6}        &\textbf{avg-pooling} &\textbf{avg-pooling} & $\ominus$&\textbf{65.5} \\\Xhline{3\arrayrulewidth}      
\end{tabular}
\label{tab:ops}
}
\vspace{-5mm}
\end{center}
\end{table}

\subsection{Point Cloud Object Detection}
\noindent \textbf{Object Detection Baselines.}
VoteNet~\cite{qi2019deep} is an end-to-end generic 3D point cloud object detection network. With the point features learned from the backbone, VoteNet generates some votes indicating the directions to the centers of candidate objects, then predicts semantic labels and 3D bounding boxes based on the votes. As a simplified version, BoxNet~\cite{qi2019deep} directly makes the predictions based on the backbone's output without such a voting process. In contrast, ImVoteNet~\cite{qi2020imvotenet} extends the pipeline of VoteNet by lifting additional 2D votes in images for 3D votes in the point cloud.

\vspace{2mm}
\noindent \textbf{Object Detection Dataset.}
SUN RGB-D~\cite{song2015sun} is a single-view RGB-D dataset, which contains 5285 samples for training and 5050 testing examples. Based on the provided camera parameters, the corresponding 3D point cloud data can be generated. To fairly compare with the results in~\cite{qi2019deep, qi2020imvotenet}, we only use the 3D coordinates as our input. Following the evaluation protocol in~\cite{qi2019deep, qi2020imvotenet}, we report the ten most common object categories in the dataset.

\vspace{2mm}
\noindent \textbf{Experimental Settings.}
As the point cloud object detection pipeline often takes a backbone to generate the seed point features, it is better to use our proposed module in the backbone rather than the voting process. In practice, since all the backbones in the three baseline networks are built with SA and FP layers of PointNet++, we also place the PnP-3D module after each SA layer as in Section~\ref{sec:cls} and~\ref{sec:seg}.

\vspace{2mm}
\noindent \textbf{Object Detection results.}
Table~\ref{tab:sunrgbd} clearly indicates the increasing margin caused by using the PnP-3D module in the baseline's backbone. Concretely, our approach benefits the categories of \emph{bathtub, bookshelf, desk, nightstand, table} in all baselines, where the highest growth is over 10\%. Although the overall improvements (by 1.4\% to 1.7\% mAP) are not as significant as the ones in Table~\ref{tab:cls} and~\ref{tab:s3dis} since only the backbone is modified, we still achieve a state-of-the-art result (63.8\% mAP) on the SUN RGB-D dataset with ImVoteNet~\cite{qi2020imvotenet} baseline. 

\begin{table}
\begin{center}
\caption{Generating the global bilinear response ($\eta_{ij}$) with point-wise response ($\lambda_{i}$) and channel-wise response ($\mu_{j}$), tested on RandLA-Net~\cite{hu2020randla} and \emph{Area 5} of \emph{S3DIS}~\cite{armeni2017joint} dataset.}
\vspace{-3mm}
\resizebox{0.92\columnwidth}{!}{
\begin{tabular}{c|c|c|c}
\Xhline{3\arrayrulewidth}
Model    & Meaning of $\eta_{ij}$ & Formula (in Equation~\ref{equ:g_mean}) &mIoU \\\hline
1 &Sum & $\lambda_{i} + \mu_{j}$&64.7\\
2 &Product & $\lambda_{i} \mu_{j}$&64.2\\
3 &Grand Mean & $(\lambda_{i} + \mu_{j})/2$&63.7\\
4 &Quadratic Mean & $\mathrm{sqrt}({{\lambda}_i}^2 + {{\mu}_j}^2)$&64.4\\
5 &Harmonic Mean & $(2\lambda_i \mu_j)/(\lambda_{i} + \mu_{j})$&65.2\\
\hline
\textbf{6}        &\textbf{Geometric Mean} & $\mathrm{sqrt}({{\lambda}_i {\mu}_j})$&\textbf{65.5} \\\Xhline{3\arrayrulewidth}      
\end{tabular}
\label{tab:global}
}
\vspace{-5mm}
\end{center}
\end{table}

\subsection{Ablation Studies}
\label{sec:abl}
\noindent \textbf{Pooling and Regularization Strategy.}
In addition to the average-pooling operation used in Equation~\ref{equ:global_c} and~\ref{equ:global_p}, alternatively, the max-pooling operation can extract the \emph{prominent} features to regularize (Equation~\ref{equ:out}) the feature map $\mathcal{F}_L$ learned from our local context fusion block. To investigate a best combination, we conduct experiments using different pooling (\ie, max/average-pooling) and regularization (\ie, element-wise product/summation/subtraction) strategies. The result of model 6 in Table~\ref{tab:ops} indicates that filtering out (\ie, subtracting) the \emph{common} features (\ie, extracted by the average-pooling operation) from $\mathcal{F}_L$ is more effective.  

\vspace{2mm}
\noindent \textbf{Global Response Generation.}
Another set-up that needs further investigation is the way of generating a global response $\eta_{ij}$ using both point-wise response $\lambda_{i}$ and channel-wise response $\mu_{j}$. Besides Equation~\ref{equ:g_mean} calculating the \emph{geometric mean} of $\lambda_{i}$ and $\mu_{j}$, we explore other possible usages in Table~\ref{tab:global}. The experimental results show that a higher-order mean value (\eg, quadratic/harmonic/geometric mean) can better estimate a global response for spatial and channel-related information, where the geometric mean (model 6) achieves the highest performance.

\vspace{2mm}
\noindent \textbf{Comparisons with Attention Modules.}
As introduced in Section~\ref{sec:intro} and~\ref{sec:work}, the attention mechanism can also be applied in point cloud networks with the intention of refining the feature map. To this end, we compare the effectiveness of the PnP-3D module with \emph{five} recent 3D attention approaches to point cloud object detection. According to the results shown in Table~\ref{tab:attention}, our approach outperforms the competitors under the mAP metric in both IoU thresholds of 0.25 and 0.5. 

In particular, we notice that some attention approaches~\cite{xie2018attentional, feng2020point, guo2021pct} cannot benefit the baseline network~\cite{qi2019deep} since they excessively exploit one type of information (\eg, point-wise) causing redundancies in the feature representations. Instead, the PnP-3D module captures both local and global context following a compact design, balancing the effectiveness and computational cost. As shown in Table~\ref{tab:attention}, our method costs the fewest parameters while achieving the highest performance. Compared to the latest Point-Transformer~\cite{zhao2020point}, our PnP-3D shows much higher efficiency \wrt the number of FLOPs. 
\begin{table}
\begin{center}
\caption{Comparisons with the attention modules using VoteNet~\cite{qi2019deep} on \emph{SUN RGB-D V1}~\cite{song2015sun} dataset.(\enquote{Dual}: from both point-axis and channel-axis; \enquote{Local}: from local areas; the value behind \enquote{@}: IoU threshold.)}
\vspace{-3mm}
\resizebox{0.95\columnwidth}{!}{
\begin{tabular}{c|c|ccc|c|c}
\Xhline{3\arrayrulewidth}
\multirow{2}{*}{Method} & Additional &Model Size &Parameters &FLOPs & \multicolumn{2}{c}{mAP (\%)} \\\cline{6-7}
&Info Origin &(MB) &$(\times10^6$) &$(\times10^9$) &@0.25&@0.5\\\hline
baseline~\cite{qi2019deep} & -- &10.9 &0.95  &48.5      & 57.7 & 33.1         \\\hline
+A-SCN~\cite{xie2018attentional} & Point-axis &15.9 &1.39  &\textbf{48.8}            & 55.6 & 30.1         \\
+Point-attn~\cite{feng2020point} & Point-axis  &15.9 &1.39  &\textbf{48.8}           & 56.4 & 32.2         \\
+CAA~\cite{qiu2021geometric} & Channel-axis  &34.7 &3.03  &50.3         & 58.8 & 33.3         \\
+Offset-attn~\cite{guo2021pct} & Point-axis &19.5 &1.73  &50.4           & 55.7 & 30.6         \\
+Point-Trans~\cite{zhao2020point}   & Dual + Local  &25.7 &2.24  &126.0        & 58.1 & 34.8         \\
\hline
\textbf{+PnP-3D} & Dual + Local  &\textbf{14.4} &\textbf{1.25} &50.2      & \textbf{59.1} & \textbf{34.9} \\\Xhline{3\arrayrulewidth}      
\end{tabular}
\label{tab:attention}
}
\vspace{-3mm}
\end{center}
\end{table}

\begin{table}
\begin{center}
\caption{Comparisons with the task-specific modules using different baselines, tested on \emph{ModelNet40}~\cite{wu20153d} (classification), \emph{S3DIS}~\cite{armeni2017joint} (semantic segmentation) and \emph{SUN RGB-D V1}~\cite{song2015sun} (object detection) datasets.}
\vspace{-3mm}
\resizebox{0.95\columnwidth}{!}{
\begin{tabular}{c|c|c|c|c}
\Xhline{3\arrayrulewidth}
\multirow{4}{*}{Classification} &Method & OA &+PA~\cite{li2020pointaugment} &+PnP-3D \\\cline{2-5} 
&PointNet++~\cite{qi2017pointnet++}&90.7&92.9&\textbf{93.2}\\\cline{2-5} 
&DGCNN~\cite{wang2019dynamic}&92.2&\textbf{93.4} &\textbf{93.4}\\\cline{2-5} 
&RS-CNN~\cite{liu2019relation}&92.2&92.7&\textbf{93.1}\\\hline 
\multirow{3}{*}{Segmentation} &method & mIoU &+CGA~\cite{lu2021cga} &+PnP-3D \\\cline{2-5} 
&RandLA-Net~\cite{qi2017pointnet++}&62.5&65.4&\textbf{65.7}\\\cline{2-5} 
&CloserLook3D~\cite{liu2020closer}&65.7&\textbf{68.6} &68.5\\\hline
\multirow{3}{*}{Detection} &method & mAP &+RGB color &+PnP-3D \\\cline{2-5} 
&VoteNet~\cite{qi2019deep}&57.7 &56.3 &\textbf{59.1}\\\cline{2-5} 
&ImVoteNet~\cite{qi2020imvotenet}&62.3 &63.4 &\textbf{63.8}\\\cline{2-5} 
\Xhline{3\arrayrulewidth}      
\end{tabular}
\label{tab:compare}
}
\vspace{-3mm}
\end{center}
\end{table}

\vspace{2mm}
\noindent \textbf{Comparisons with Task-Specific Modules.}
We also compare our approach with different task-specific modules in all three tasks. As Table~\ref{tab:compare} indicates, the PnP-3D module provides higher improvements than the Point-Augment~\cite{li2020pointaugment} framework in classification; while in semantic segmentation, we achieve the comparable results as using the CGA~\cite{lu2021cga} module. Although there is no such a plug-and-play module for object detection, our approach can better benefit the baselines than feeding additional color information. Overall, our proposed plug-and-play module is more effective and generic for point cloud analysis tasks.   

\subsection{Visualizations}
\begin{figure}
\begin{center}
\includegraphics[width=0.95\columnwidth]{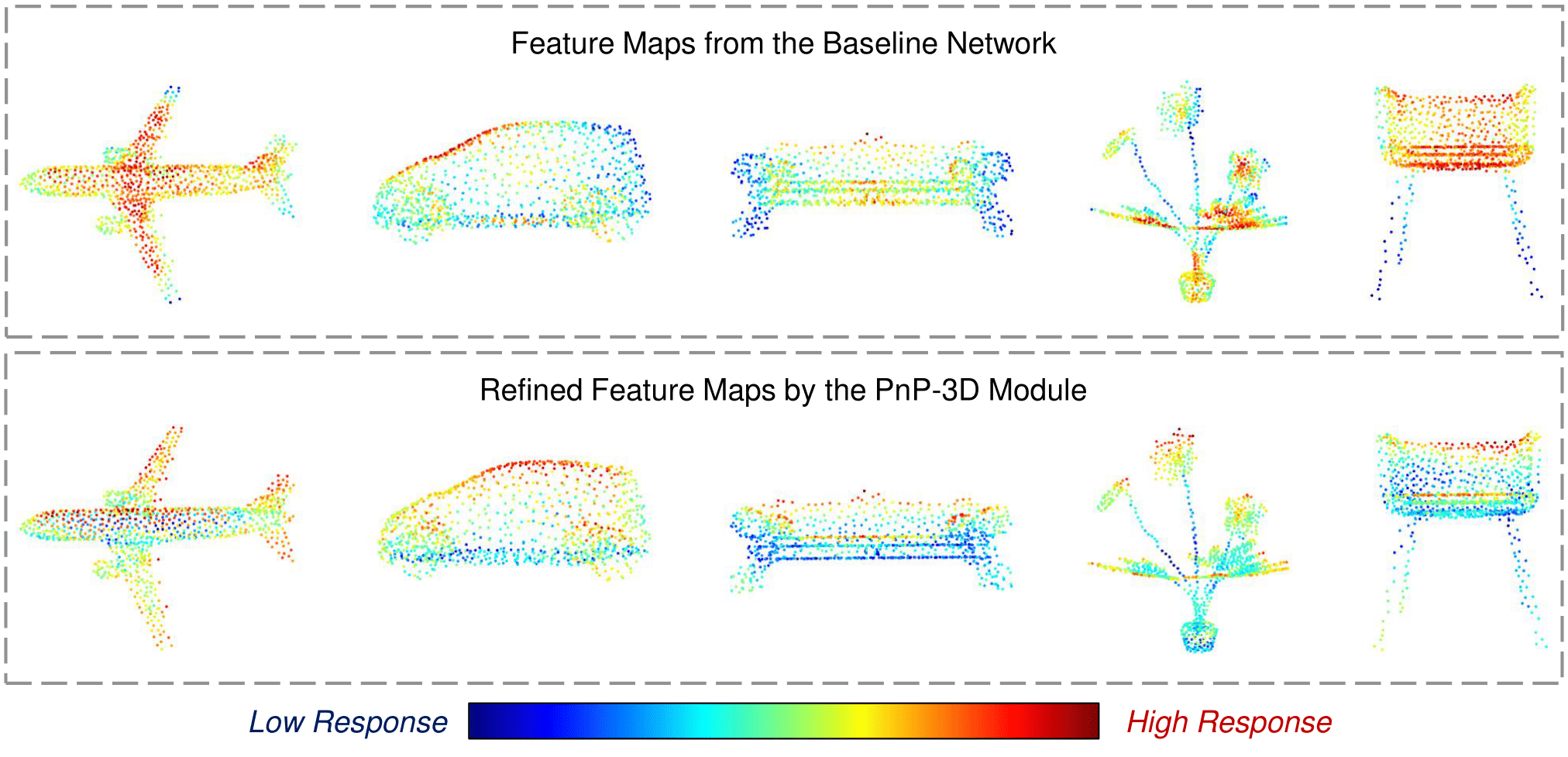}
\end{center}
\vspace{-3mm}
\caption{Visualization of the feature maps in \emph{ModelNet40}~\cite{wu20153d} classification. The first row shows the features learned from DGCNN~\cite{wang2019dynamic}, while the second row is the refined output from the PnP-3D module. We normalize the channels for a heat-map view. It can be clearly observed that our module better illustrates the outlines of point cloud objects.}
\label{fig:vis2}
\end{figure}

To analyze the PnP-3D module's behavior, we visualize and compare the feature maps in Figure~\ref{fig:vis2}. Generally, we observe that our approach can raise the high responses in more representative areas (\eg, the wings/tail of plane, the armrest/leg of sofa/chair) covering a complete outline of point cloud object, while the baseline method (DGCNN~\cite{wang2019dynamic}) only focuses on the central parts. This advantage can be credited to the global bilinear regularization block, which sharpens the feature map by integrating more global information based on both spatial and channel-related clues. Further, this property also benefits the semantic segmentation task as shown in Figure~\ref{fig:vis1}, where the number of \enquote{confusing} points (\ie, near the boundaries of different categories) has been remarkably reduced. 

\begin{figure}
\begin{center}
\includegraphics[width=0.95\columnwidth]{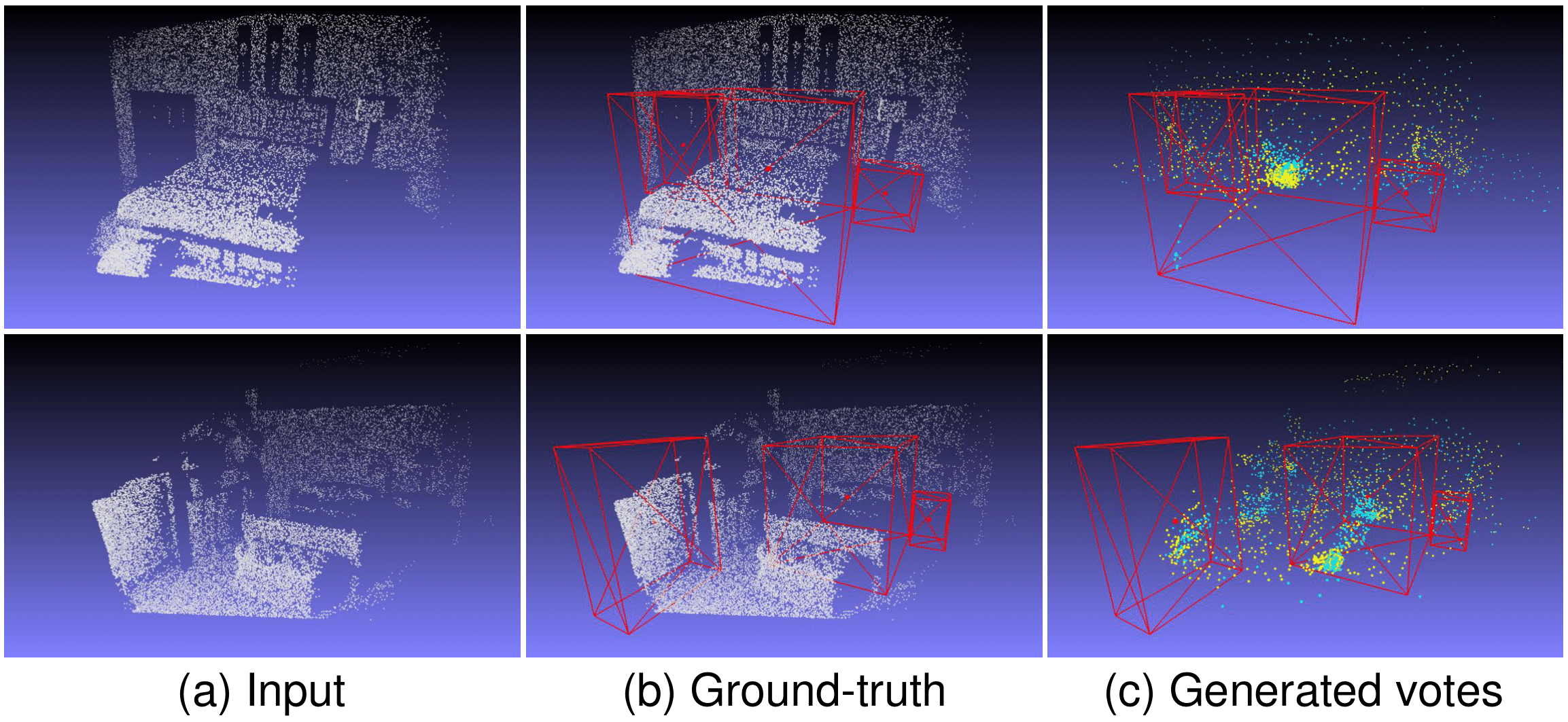}
\end{center}
\vspace{-3mm}
\caption{Examples of the generated votes in \emph{SUN RGB-D}~\cite{song2015sun} object detection. The bounding boxes (ground-truth) are in red frames, where the red points are the centroids. In the last column, we can find that the votes generated by the PnP-3D (blue points) better approach the object centroids than the baseline's outputs (VoteNet~\cite{qi2019deep}, yellow points).}
\label{fig:vis3}
\vspace{-3mm}
\end{figure}

In addition, we compare the generated votes for object detection in Figure~\ref{fig:vis3}. Recall that in the VoteNet~\cite{qi2019deep} baseline, the votes are generated from the backbone's output feature map to estimate the centroids of detected objects. By leveraging the PnP-3D module in the backbone, more votes can be closely attached to the centroids (\eg, shown as the right object in the top-right subfigure and the middle object in the bottom-right subfigure of Figure~\ref{fig:vis3}), providing more confident estimations for object proposals. 

\section{Conclusion}
In this paper, we focus on how to better refine the feature representations of point cloud data with a simple \emph{plug-and-play}. To address this fundamental problem in point cloud analysis, we propose an effective module named PnP-3D, consisting of two specifically designed blocks. Concretely, the local context fusion block can fuse both local geometric and feature context according to the inherent point distribution in 3D space. Moreover, the global bilinear regularization block is leveraged to regularize the features by aggregating the global responses based on both point-wise and channel-wise information in feature space. By utilizing our module in different baselines and various point cloud analysis tasks, we comprehensively demonstrate the effectiveness of the PnP-3D module. The ablation studies and visualizations further verify the properties and behavior of our approach. In the future, we expect to extend its usage in low-level vision tasks such as point cloud upsampling or completion, and optimize its efficiency for real-time applications.   

\bibliographystyle{IEEEtran}
\bibliography{egbib}

\end{document}